\documentclass[conference]{IEEEtran}
\usepackage{xspace}

\IEEEoverridecommandlockouts
\usepackage{cite}
\usepackage{amsmath,amssymb,amsfonts}
\usepackage{algorithmic}
\usepackage{graphicx}
\usepackage{textcomp}
\usepackage{xcolor}
\usepackage{lipsum}
\usepackage{duckuments}
\usepackage{hyperref}
\usepackage[table,dvipsnames]{xcolor}
\usepackage{multirow}
\usepackage{adjustbox}
\usepackage{booktabs}
\usepackage{tabularx}
\usepackage{comment}

\def\BibTeX{{\rm B\kern-.05em{\sc i\kern-.025em b}\kern-.08em
    T\kern-.1667em\lower.7ex\hbox{E}\kern-.125emX}}
\begin{document}

\newcommand{\temp}[1]{\textcolor{blue}{#1}}
\newcommand{\methodname}{Zero-shot Decentralized Federated Learning\xspace}
\newcommand{\methnam}{ZeroDFL\xspace}
\newcommand{\zerogray}{\textcolor{gray!10}{0}}
\newcommand{\zerowhite}{\textcolor{white}{0}}
\newcommand{\cumulativeourbest}{807~\text{MB}\xspace}
\newcommand{\cumulativeournormal}{4~\text{GB}\xspace}
\newcommand{\cumulativeourworst}{46.48~\text{GB}\xspace}
\newcommand{\cumulativefedtpg}{467~\text{GB}\xspace}
\newcommand{\reductionourbest}{592×\xspace}
\newcommand{\reductionournormal}{118×\xspace}
\newcommand{\reductionourworst}{10×\xspace}

\title{Zero-shot Decentralized Federated Learning}

\author{\IEEEauthorblockN{Alessio Masano, Matteo Pennisi, Federica Proietto Salanitri, Concetto Spampinato, Giovanni Bellitto }
\IEEEauthorblockA{PeRCeiVe Lab, University of Catania, Italy.\\ 
Email: alessio.masano@phd.unict.it; \{name.surname\}@unict.it}
}

% \author{\IEEEauthorblockN{}}

\maketitle

\begin{abstract}
CLIP has revolutionized zero-shot learning by enabling task generalization without fine-tuning. While prompting techniques like CoOp and CoCoOp enhance CLIP’s adaptability, their effectiveness in Federated Learning (FL) remains an open challenge. Existing federated prompt learning approaches, such as FedCoOp and FedTPG, improve performance but face generalization issues, high communication costs, and reliance on a central server, limiting scalability and privacy.

We propose \methodname (\textbf{\methnam}), a fully decentralized framework that enables zero-shot adaptation across distributed clients without a central coordinator. \methnam employs an iterative prompt-sharing mechanism, allowing clients to optimize and exchange textual prompts to enhance generalization while drastically reducing communication overhead.

We validate \methnam on nine diverse image classification datasets, demonstrating that it consistently outperforms—or remains on par with—state-of-the-art federated prompt learning methods. More importantly, \methnam achieves this performance in a fully decentralized setting while reducing communication overhead by 118× compared to FedTPG. These results highlight that our approach not only enhances generalization in federated zero-shot learning but also improves scalability, efficiency, and privacy preservation—paving the way for decentralized adaptation of large vision-language models in real-world applications. Code is available at: \url{https://github.com/perceivelab/ZeroDFL}
\end{abstract}

\section{Introduction}
A key breakthrough in vision-language research has been CLIP (Contrastive Language-Image Pre-training), which has revolutionized zero-shot learning by enabling models to generalize to new tasks without requiring task-specific fine-tuning~\cite{clip}. By aligning images and text within a shared feature space, CLIP learns rich, highly transferable representations that allow it to perform a wide range of vision-language tasks with remarkable flexibility. This capability has positioned CLIP as a foundational model for applications where adaptability and generalization across diverse domains are critical.

However, fine-tuning CLIP on specific application domains is often infeasible due to the high complexity of the model and constrained computational resources. This limitation is particularly evident when dealing with highly specialized domains, such as medical imaging, or when data availability is extremely constrained. Fine-tuning such large models, initially trained on hundreds of millions of text-image pairs, would require large computational power and large-scale datasets, which are often unavailable.

To overcome this challenge, prompting has emerged as a viable alternative for adapting CLIP to specific domains. Prompting techniques were first introduced in the context of natural language processing (NLP), where they were used to guide pre-trained models toward specific tasks without modifying their internal weights~\cite{lester2021power}. In the context of CLIP, prompting has been applied mainly in two ways: visual prompts and text prompts. Visual Prompt Tuning (VPT) involves prepending learnable visual tokens to the input of CLIP’s visual encoder~\cite{vpt}, while textual prompting methods such as CoOp~\cite{coop} and CoCoOp~\cite{cocoop} prepend learnable text embeddings to CLIP's text encoder. These methods have proven effective in enhancing CLIP’s zero-shot learning capabilities without requiring full fine-tuning. 
%However, while CLIP has demonstrated strong performance in centralized training settings, its effectiveness in distributed settings like Federated Learning (FL) remains an open challenge. 
%FL is a learning paradigm that enables multiple clients to collaboratively train a model without sharing their original training data, thereby preserving privacy.
However, while CLIP has demonstrated strong performance in centralized training environments, its effectiveness in distributed scenarios, such as Federated Learning (FL), remains an open challenge. Unlike traditional centralized training, where data is aggregated in a single location, FL is a learning paradigm that enables multiple clients to collaboratively train a model while keeping their raw data localized. This approach not only enhances privacy by preventing data exposure but also poses unique challenges, such as communication overhead, data heterogeneity, and model convergence, especially for zero-shot learning.

Incorporating prompting strategies into FL presents a promising direction for enabling zero-shot classification also in federated settings. Both visual and text-based prompts can facilitate adaptation across clients, but text-based prompting offers a crucial advantage: it is inherently more resilient to adversarial attacks compared to visual prompting. This resilience enhances privacy preservation, a critical requirement in federated systems~\cite{fowl2022decepticons}. FedCoOp~\cite{promptfl} integrates the CoOp approach into FL, improving CLIP’s performance on seen classes within each client. However, a major limitation arises: prompt vectors optimized on local data often fail to generalize to unseen classes—whether from other clients or newly emerging categories—necessitating additional adaptation mechanisms. To address this, FedTPG~\cite{fedtpg} introduces a local prompt generator, allowing each client to optimize its own generator independently before sharing it with a central server, in order to construct a global model. While this approach enhances generalization across unseen classes compared to FedCoOp, its reliance on a central server introduces drawbacks, including a single point of failure and communication bottlenecks.

To address these challenges, we propose a simple yet effective approach, \textbf{\methodname (\methnam)}, which enables fully decentralized zero-shot classification by learning and exchanging text prompts. Unlike traditional federated methods that rely on a central server, \methnam operates in a completely decentralized manner, where clients independently optimize and share prompt representations without any centralized coordination.

%\methnam
ZeroDFL’s optimization follows a two-step process. First, each client performs local training using its own prompt vectors. Once optimization is complete, it randomly shares its prompt set with other clients in the federation. Upon receiving external prompts, each node selects a subset of the received prompts and adapts them using its private data. This iterative exchange and adaptation process continues across multiple training rounds, allowing all clients to collaboratively refine their prompt representations.

Despite operating in a more challenging, fully decentralized setting, \methnam achieves generalization performance that is on par with or even superior to centralized approaches as demonstrated by an extensive evaluation carried out on on nine diverse image classification datasets, including Caltech101, Flowers102, Oxford Pets, FGVC Aircraft, DTD, UCF101, Food101, Stanford Cars, and SUN397. Our method not only yields satisfactory zero-shot classification accuracy, but, more importantly, achieves this performance in a fully decentralized setting while significantly reducing communication costs, showing, for instance, a \textbf{\reductionournormal} reduction in transmitted data compared to (our main competitor) FedTPG~\cite{fedtpg}.

These results cumulatively highlight that our decentralized approach achieves state-of-the-art performance while enhancing scalability, efficiency, and privacy preservation in federated zero-shot learning.

\section{Related Work}

Prompting techniques for Vision-Language Models (VLMs) have shown promising results in downstream image classification tasks~\cite{clip,align,convirt}. A significant advancement in this field was introduced by CoOp \cite{coop}, which proposed Contextual Optimization, an approach where contextual vectors replace hand-crafted prompts. These context vectors are concatenated with the image label to form the text encoder input of a pre-trained VLM. By keeping the VLM frozen, the context vectors are optimized for the specific task.
A further improvement was introduced by CoCoOp \cite{cocoop}, where context vectors are conditioned on image features through a neural network. Other approaches explore visual rather than textual prompts. For instance, \cite{vpt} injects learnable prompts into the input layer of the vision transformer alongside the CLS token and patch embeddings. In \cite{upt}, a mixed approach is proposed, where a small neural network is trained to jointly optimize prompts across different modalities. Similarly, \cite{maple} presents a joint prompting strategy, where context prompts are learned in both vision and language branches, with textual prompts conditioning visual prompts via a coupling function. To enable hierarchical representation learning, prompts are integrated at multiple depths across different transformer blocks.

Federated Learning (FL) enables multiple clients to collaboratively train models without sharing raw data, making it a privacy-preserving alternative to centralized learning~\cite{mcmahan2017communication,kairouz2021advances}. A decentralized variant of FL eliminates the need for a central server, allowing direct client-to-client communication, reducing coordination overhead, and improving scalability~\cite{10542323}. However, standard FL approaches often involve full model training and aggregation, leading to high communication costs and computational overhead.

To address these challenges, prompt-based FL has emerged as an efficient alternative. Instead of transmitting entire models, clients exchange soft prompts, significantly reducing communication costs while maintaining adaptability. FedCoOp\cite{promptfl} first introduced this concept by enabling federated clients to train and share textual prompts. pFedPG\cite{pfedpg} improved adaptability by introducing a server-side prompt generator to personalize prompts for each client, ensuring better alignment with heterogeneous local data. FedTPG\cite{fedtpg} refined this approach further by employing a text-conditioned local prompt generator, which enhances generalization to unseen data by adapting prompts to local distributions before aggregation. FedPGP\cite{fedpgp} introduced a low-rank adaptation mechanism to balance global generalization and local personalization.

Despite these advancements, existing methods rely on centralized coordination, either for model aggregation or prompt personalization, introducing bottlenecks, scalability issues, and potential failure points. In contrast, \methnam is fully decentralized, eliminating the need for a central authority. Clients independently optimize, exchange, and refine textual prompts in a peer-to-peer fashion, iteratively improving zero-shot capabilities across distributed environments.

\section{Method}
\begin{figure*}[htb!]
    \centering
    \includegraphics[width=1\textwidth]{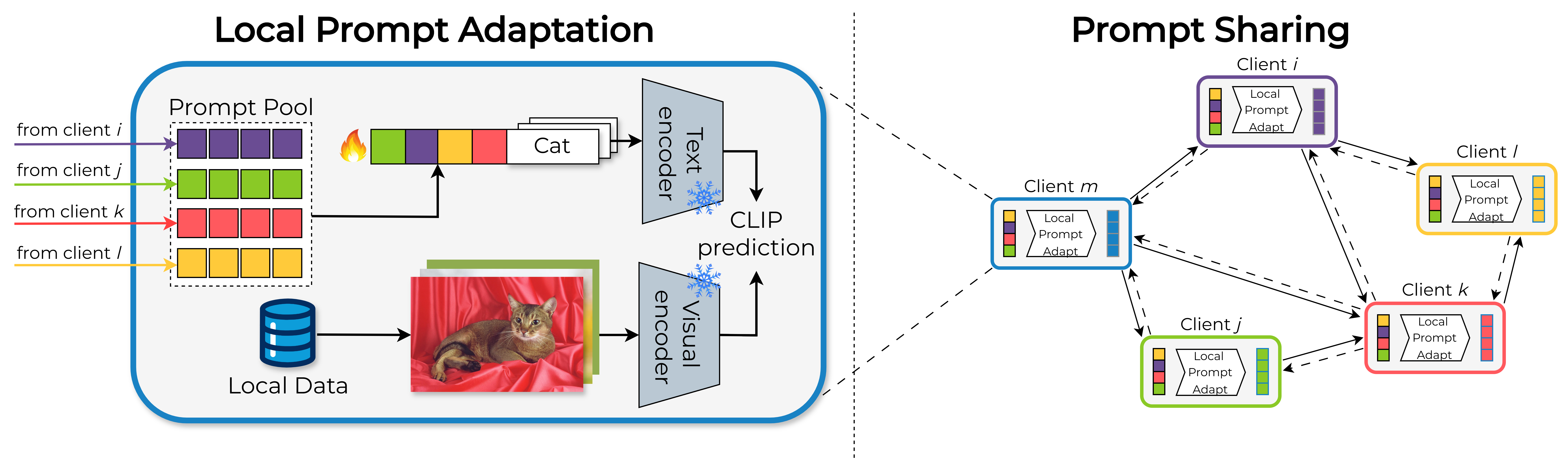}
    %\caption{\textbf{An overview of our proposed method}. Left side: At the beginning of a new round, each client randomly selects $M$ textual prompt from the Prompt Pool received from other clients in the federation and performs local adaptation using its own set of local classes. Right side: Once the optimization is completed, the refined prompts are sent to a subset of federation clients, randomly chosen in each round.}
    \caption{\textbf{\methnam strategy overview}. \methnam operates in iterative training rounds, each consisting of two phases: local adaptation and prompt exchange. \textit{Left: Local Adaptation} – Each client selects $M$ textual prompts from its Prompt Pool (prompts received from other clients) and fine-tunes them by prepending them to a frozen CLIP text encoder and optimizing on its private dataset. \textit{Right: Prompt Exchange} – After local adaptation, the client shares its updated prompts with S selected clients, prioritizing those that have received fewer updates in previous rounds, ensuring balanced knowledge distribution.%This adaptive exchange strategy promotes a more balanced and efficient dissemination of knowledge in a fully decentralized manner.
    }
    \label{fig:method}
\end{figure*}

\methnam is fully decentralized learning approach that enables clients to collaboratively refine knowledge without relying on a central authority. In our setup, a group of nodes, referred to as \emph{clients}, form a \emph{federation}, where each client maintains a private classification dataset with a non-IID distribution and communicates directly with other clients to exchange knowledge.

As illustrated in Fig.~\ref{fig:method}, learning proceeds in iterative \emph{rounds}, where each round consists of two key phases: \emph{local adaptation} and \emph{prompt exchange}. At the beginning of each round (Fig.\ref{fig:method}, \textit{Left}), each client selects a subset of $M$ textual prompts from a \emph{prompt pool}, which contains prompts received by other clients. These prompts are then refined locally using the client’s private dataset, allowing adaptation to its specific data distribution. Once this optimization step is complete (Fig.\ref{fig:method}, \textit{Right}), the updated prompt vectors are sent to a set of $S$ selected clients. To ensure fair knowledge distribution across the federation, the recipient selection mechanism prioritizes $S$ clients that have received fewer updates in previous rounds.

\subsection{Local adaption through prompt learning}

Let's now formalize the \methnam's local adaption strategy. Let \( F \) be a federation of \( C \) clients, where each client \( c_i \in C\) maintains a private dataset \( \mathcal{D}_i \). Each dataset consists of pairs \( (\mathbf{x}, y) \sim \mathcal{D}_i \), where \( \mathbf{x} \in \mathcal{X}_i \) is an input sample (e.g., an image), and \( y \in \mathcal{Y}_i \) its corresponding label. The datasets across the \( C \) clients follow a \textit{non-IID} distribution, meaning that class sets are disjoint, i.e., \( \mathcal{Y}_i \cap \mathcal{Y}_h = \emptyset \) for \( i \neq h \).  

Each client \( c_i \) also has access to the textual names of its target classes, denoted as \( \left\{ t_{i,j} \right\}_{j=1}^{Q_i} \), where \( Q_i = |\mathcal{Y}_i| \) is the number of unique classes in \( \mathcal{D}_i \). 

CLIP~\cite{clip} is a \textit{vision-language model} (VLM) consisting of two encoders: a \textit{visual encoder} \( f(\cdot) \) and a \textit{text encoder} \( g(\cdot) \), trained through contrastive learning. Given an input image \( \mathbf{x} \), CLIP embeds it as \( f(\mathbf{x}) \), while the text encoder maps a textual description \( t \) into the shared space as \( g(\varepsilon(t)) \), where \( \varepsilon(\cdot) \) denotes the tokenizer that converts text into numeric tokens.

Since fine-tuning the entire CLIP model is computationally expensive, we employ \textit{prompt learning}, i.e., instead of updating the entire CLIP's parameters, a set of \textit{learnable textual prompt vectors} is prepended to class names in the text encoder, adapting the model to specific tasks.

Thus, for each client \( c_i \), we introduce a set of \( M \) learnable prompt vectors:

\[
V_i = \left\{ v_{i,1}, v_{i,2}, \dots, v_{i,M} \right\},
\]

where each \( v_{i,m} \) is a trainable vector of dimension \( d \), matching the size of the CLIP text embeddings. These prompts are then prepended to the encoding of the class names. 
For a given class \( q \), the \textit{augmented prompt representation} input to the CLIP text encoder \( g(\cdot) \) is defined as:  

\[
P_q = \left[ v_{i,1}, v_{i,2}, \dots, v_{i,M}, \varepsilon(t_{i,q}) \right].
\]

Here, \( V_i \) is learned at the client level, meaning that the same set of prompt vectors is shared across all classes within client \( c_i \). In contrast, \( P_q \) represents the final class-specific text input, where the learnable prompts are prepended to the emmbedded class name $t_{i,q}$ .

The probability of predicting class \( q \) for a given input image \( \mathbf{x} \) is computed using a softmax function over the similarity scores between the image embedding and each class-specific prompt representation:

\[
p(\hat{y} = q \mid \mathbf{x}) =
\frac{\exp(\phi(f(\mathbf{x}), g(P_q)) / \tau)}
{\sum_{h=1}^{Q_i} \exp(\phi(f(\mathbf{x}), g(P_h)) / \tau)},
\]

where \( \phi(\cdot, \cdot) \) denotes a similarity function (e.g., cosine similarity), \( \hat{y} \) is the predicted label, \( Q_i \) is the number of classes in client \( c_i \), and \( \tau \) is the softmax temperature parameter.

During training, we optimize the learnable prompts in \( V_i \) to minimize the negative log-likelihood loss:

\[
\mathcal{L} = - \mathbb{E}_{\mathbf{x}, y \sim \mathcal{D}_i} \left[ \log p(y \mid \mathbf{x}) \right].
\]

\subsection{Prompt Exchange Strategy}

Unlike previous methods that rely on a central server to aggregate client updates before distributing them across the federation, \methnam enables a fully decentralized peer-to-peer prompt exchange. After local adaption, described in the previous section, each client selects a subset of \( S \) other clients within the federation to exchange its own prompts with. We employ a \textit{weighted selection strategy} that prioritizes clients who have been chosen less frequently in previous rounds. This mechanism ensures that knowledge is progressively and evenly distributed across the federation.

Each client \( c_i \) maintains a history of selection frequencies for every other client in the federation. Specifically, the \textit{selection frequency} \( F_{i,j}^r \) denotes the number of times client \( c_j \) has been selected as a recipient by client \( c_i \) up to round \( r \). To determine the recipients for prompt exchange, each client \( c_i \) assigns a weight to every other client \( c_j \)  based on the inverse of its selection frequency:

\begin{equation}
    w_{i,j}^r = \frac{1}{F_{i,j}^r + \epsilon}
\end{equation}

where \( \epsilon \) is a small constant (e.g., \( 10^{-6} \)) to prevent division by zero. This formulation ensures that clients who have been selected less frequently by \( c_i \) are given higher priority in subsequent rounds.

The final selection of recipient clients is performed by sampling \( S \) clients according to a probability distribution derived from these weights. The probability of selecting client \( c_j \) as a recipient  by client \( c_i \) is then computed as:

\begin{equation}
\pi_i(c_j) = \frac{w_{i,j}^r}{\sum_{k \neq i} w_{i,k}^r}.
\end{equation}

This strategy promotes an even distribution of knowledge across the federation, ensuring that prompt exchanges are balanced over multiple rounds.

Once the prompt exchange phase is completed, each client \( c_j \) receives prompt vectors from a variable number of other clients, forming its \textit{prompt pool} \( \mathcal{P}_j^r \). To maximize the diversity of received knowledge, each client then samples \( M \) prompt vectors following a strategy that, whenever possible, ensures each selected prompt originates from a different client. This promotes diversity in the learned representations and prevents over-reliance on a small subset of sources.

Each client then performs local \textit{prompt adaptation} for \( E \) epochs, updating its prompt vectors using its private dataset before engaging in the next exchange cycle. This iterative process of \textit{adaptation and exchange} continues for \( R \) rounds, ensuring that the prompt representations gradually improve while knowledge is efficiently propagated throughout the federation.

\section{Experimental Results}

\subsection{Experimental Setup}

Our experimental setting is specifically designed to assess the \textit{zero-shot learning capabilities} of \methnam in a federated learning scenario. Our evaluation thus emphasizes \textit{generalization to unseen classes} across heterogeneous clients. To this end, we carefully select diverse datasets, compare against state-of-the-art federated prompt learning approaches, and design training scenarios that effectively capture the challenges of federated zero-shot classification.

\subsubsection{\textbf{Datasets}}
Following the setup in \cite{fedtpg}, we use nine image classification datasets covering a broad range of visual recognition tasks.  
\textit{Caltech101}~\cite{caltech} (101 classes) consists of images from diverse object categories.  
\textit{Flowers102}~\cite{flowers} (102 classes) contains different species of flowers.  
\textit{OxfordPets}~\cite{pets} (37 classes) includes various breeds of cats and dogs.  
\textit{FGVCAircraft}~\cite{fgvcaircraft} (100 classes) is a fine-grained dataset for aircraft recognition.  
\textit{DTD}~\cite{dtd} (47 classes) comprises images of textural patterns.  
\textit{UCF101}~\cite{ucf101} (101 classes) consists of frames extracted from action recognition videos.  
\textit{Food101}~\cite{food101} (101 classes) contains images of various food dishes.  
\textit{StanfordCars}~\cite{stanfordcars} (196 classes) focuses on distinguishing different car models and makes.  
\textit{SUN397}~\cite{sun397} (397 classes) is a large-scale dataset for scene recognition.  

These datasets collectively provide a robust benchmark for evaluating federated zero-shot learning under both \textit{heterogeneous} and \textit{homogeneous} data distributions.

\subsubsection{\textbf{Federated Learning Setup}}

To assess generalization, we split the classes of each dataset into two groups: half are used for training, while the remaining half are reserved for testing, ensuring that models must classify \textit{unseen categories} during inference. The training classes are disjointedly assigned to clients based on the chosen number of class per client \(N\).  Under this setup, local training at each client follows an \textit{\(N\)-way, \(K\)-shot} learning problem, being \(K\) the number of training images per class.

%To assess generalization, we split the classes of each local dataset into two groups: half are used for training, while the remaining half are reserved for testing, ensuring that models must classify \textit{unseen categories} during inference. Under this setup, local training at each client follows an \textit{\(N\)-way, \(K\)-shot} learning problem, where each client operates on \(N\) disjoint classes from its assigned dataset, with \(K\) training images per class.

We simulate two \textit{federated learning scenarios}. In the \textit{heterogeneous scenario}, we construct a federation where each client operates on a completely different dataset, following the approach in \cite{fedtpg}. This setup reflects a highly \textit{non-IID data distribution}, where the global knowledge must be shared across  different visual domains. In the \textit{homogeneous scenario}, we partition a single dataset among multiple clients, ensuring that all clients work with data from the same domain but with disjoint class subsets. %This scenario aligns with standard federated learning literature and serves as a controlled setting to evaluate the impact of decentralization.

\subsubsection{\textbf{Competitors}}
We compare \textit{\methnam} against several state-of-the-art federated prompt learning methods.  
\textit{FedCoOp}~\cite{promptfl} learns a unified prompt by averaging client-specific prompts.  
\textit{FedCoCoOp}, a federated adaptation of CoCoOp~\cite{cocoop}, conditions prompts on image features.  
\textit{FedMaple}, a federated version of Maple~\cite{maple}, optimizes prompts jointly for both text and image encoders.  
\textit{FedTPG}~\cite{fedtpg} trains a local text-conditioned prompt generator that adapts prompts to each client's data distribution.  

All these centralized federated methods employ FedAvg~\cite{fedavg} as their aggregation strategy.

Since these competing methods rely on a centralized server for model aggregation, while our approach operates in a fully decentralized manner, we report \textit{the average accuracy across individual clients} to ensure a fair comparison.

\subsubsection{\textbf{Implementation Details}}
All methods leverage a frozen CLIP model with ViT-B/16 as the backbone, ensuring that only the prompt vectors are optimized during training. The number of \textit{learnable prompt vectors} per client is set to \( M = 4 \), each with an embedding dimension of \( d = 512 \). Federated training is conducted for \( R = 500 \) rounds, with each client performing \textit{\( E = 1 \) local epoch per round}. Unless otherwise stated, all experiments use a \textit{\( K = 8 \), \( N = 20 \) setup}, where each client has access to 20 disjoint classes and 8 images per class for training.

\subsection{Zero-Shot Classification Performance}

We begin by evaluating the zero-shot classification performance of \methnam in the \emph{heterogeneous} setting, where data heterogeneity occurs at two levels: (i) class distribution heterogeneity, as each client possesses \( N \) non-overlapping classes, and (ii) dataset-level heterogeneity, as clients operate on entirely different datasets, leading to substantial shifts in feature distributions.

Table~\ref{tab:heterogeneous} shows that \methnam, despite being a fully decentralized approach, achieves state-of-the-art performance, yielding accuracy on par with or even superior to strong centralized approaches. In particular, \methnam achieves the highest average accuracy across all datasets (\textbf{76.19\%}), surpassing all competitors. It also outperforms all methods on Flowers102, UCF101, and Stanford Cars, highlighting its robustness in both fine-grained and large-scale classification tasks. Furthermore, it matches FedTPG's performance on Caltech101, FGVC Aircraft, and DTD, while consistently outperforming all other baselines.

\begin{table*}[htb!]
    \centering
    \renewcommand{\arraystretch}{1.2}
    \rowcolors{2}{gray!10}{white}
    \caption{Zero-shot classification Accuracy (\%) in the federated heterogeneous setting1     with $N$=20 classes per client and $C$=30 clients.}
    \label{tab:heterogeneous}
    \begin{tabularx}{\textwidth}{l|*{9}{>{\centering\arraybackslash}X}|c}
    \toprule
    \rowcolor{white}
         Method & Caltech101 & Flowers102 & Oxford Pets & FGVC Aircraft & DTD & UCF101 & Food101 & Stanford Cars & SUN397 & Avg. \\
    \midrule
           FedCoOp & 92.79 & 70.12 & 95.39 & 30.95 & 52.74 & 70.60 & 88.47 & 72.32 & 71.89 & 71.70 \\
           FedCoCoOp & 91.59 & 65.63 & 87.36 & 22.67 & 45.77 & 64.25 & 84.95 & 66.40 & 65.36 & 66.00 \\
           FedMaple & 90.06 & 68.51 & \textbf{97.09} & 32.34 & 46.61 & 68.68 & 89.60 & 71.33 & 71.33 & 70.61 \\
           FedTPG & \textbf{95.24} & \underline{77.76} & 95.79 & \textbf{35.18} & \textbf{60.51} & \underline{76.64} & \textbf{91.64} & \underline{74.26} & \textbf{77.13} & \underline{76.02} \\
        \midrule
           \methnam & \underline{94.70} & \textbf{78.67} & \underline{96.75} & \underline{34.94} & \underline{59.54} & \textbf{77.44} & \underline{91.50} & \textbf{75.38} & \underline{76.83} & \textbf{76.19} \\
         
          \bottomrule
    \end{tabularx}
\end{table*}

These results highlight the effectiveness of our strategy in achieving high zero-shot classification accuracy without requiring centralized aggregation. %The strong performance of \methnam, despite operating under a more constrained and challenging decentralized paradigm, underscores its potential as a scalable and communication-efficient solution for federated zero-shot learning.
We further examine the scalability of \methnam by analyzing its performance as the number of clients in the federation increases.
Specifically, we reduce the number of classes per client, effectively increasing the total number of clients in the federation, and assess whether this change affects model performance. Table~\ref{tab:moreclients_transposed} reports the results for $N = 10$ classes per client (thus, a federation with $C=59$ clients) in the \emph{heterogeneous} scenario. Compared to the results in Table~\ref{tab:heterogeneous}, increasing the number of clients leads to a slight decline in accuracy (about 1 percent point less than with $C=30$), due to the increased difficulty of knowledge propagation in a larger federation.  Despite this challenge, \methnam continues to perform competitively, achieving the highest average accuracy (\textbf{74.94\%}) and outperforming FedTPG on the majority of datasets and demonstrating its ability to effectively maintain model agreement even as the federation scales. 

\begin{table}[htb!]
    \centering
    \renewcommand{\arraystretch}{1.2}
    \rowcolors{2}{gray!10}{white} % Alternanza colori dalla seconda riga
    %\caption{\textbf{Zero-shot classification accuracy (\%) in the Heterogeneous scenario with increased number of clients ($N$=10 classes per client; $C$=59 clients).}}
\caption{Scalability evaluation of \methnam in the heterogeneous federated setting, where the number of clients in the federation increases. Zero-shot classification accuracy (\%) is reported for $N=10$ classes per client and $C=59$ clients.}
    \label{tab:moreclients_transposed}
    %\begin{tabularx}{\columnwidth}{l|>{\centering\arraybackslash}X>{\centering\arraybackslash}X}
    \begin{tabular}{l|cc}
    \toprule
    \rowcolor{white} 
         Dataset & FedTPG & \methnam \\
    \midrule
         Caltech101 &  94.05 & \textbf{94.47} \\
         Flowers102 &  77.30 &  \textbf{77.77} \\
         Oxford Pets &  \textbf{95.22} &  92.87 \\
         FGVC Aircraft & 33.71 &  \textbf{36.54} \\
         DTD &  56.26 &  \textbf{57.00} \\
         UCF101 & 74.35 & \textbf{74.91} \\
         Food101 &  90.72 &  \textbf{91.29} \\
         Stanford Cars &  \textbf{75.34} &  74.40 \\
         SUN397 &  \textbf{75.62} & 75.24 \\
    \midrule
         Avg. & 74.73 &  \textbf{74.94} \\
    \bottomrule
    \end{tabular}
\end{table}

These results highlight the robustness of our decentralized approach in handling larger federations without relying on a central aggregation mechanism.

Given that \methnam operates in a fully decentralized setting, it does not produce a single global model at the end of federated training, unlike centralized approaches. Instead, each client maintains its own model, resulting in multiple (as many as the number of the federation clients) independently trained models across the federation. %In Tables~\ref{tab:heterogeneous} and \ref{tab:moreclients_transposed}, we report the mean accuracy across all clients in the federation. 

To further analyze the consistency of the learned models, we conduct a convergence study, assessing how closely individual client models align in terms of performance. Specifically, for each dataset, we evaluate every client’s classifier on the global test data and compute the average accuracy. For example, in the case of Caltech101, if the dataset is partitioned among five clients, each client trains a model on its own subset and is then evaluated on the full set of unseen classes belonging to the test partition. This ensures that the reported accuracy reflects not only individual model performance but also how well knowledge is shared among clients. Table~\ref{tab:convergence} presents the number of clients and the standard deviation of accuracy for each dataset under both configurations, with $N=10$ and $N=20$. 

%To further analyze the consistency of the learned models, we conduct a convergence study, assessing how closely individual client models align in terms of performance. Specifically, for each dataset, we evaluate every client’s classifier on the data partitions of all other clients belonging to the same dataset and compute the average accuracy. For example, in the case of Caltech101, if the dataset is partitioned among five clients, each client trains a model on its own subset and is then evaluated on the test partitions of the other four clients. This ensures that the reported accuracy reflects not only individual model performance but also how well knowledge is shared among clients. Table~\ref{tab:convergence} presents the number of clients and the standard deviation of accuracy for each dataset under both configurations, with $N=10$ and $N=20$. 
The results indicate that, despite the absence of a centralized aggregation step, individual client models converge to similar performance levels as suggested by the low standard deviation across clients. This confirms two key advantages of our approach: it achieves performance comparable to centralized methods without relying on a central server for aggregation and ensures stable model consistency even as the number of clients increases.
\begin{table}[htb!]
    \centering
    \renewcommand{\arraystretch}{1.2}
    \rowcolors{2}{gray!10}{white} 
    \caption{Node convergence analysis. Standard deviation of accuracy (\%) computed across clients that share classes from the same dataset, measuring the consistency of learned models in the federation.}
    \label{tab:convergence}
    %\begin{tabularx}{\columnwidth}{l|>{\centering\arraybackslash}X>{\centering\arraybackslash}X|>{\centering\arraybackslash}X>
    %{\centering\arraybackslash}X}
    \begin{tabular}{l|cc|cc}
    \toprule
    \rowcolor{white} 
         & \multicolumn{2}{c|}{$N=10$} & \multicolumn{2}{c}{$N=20$} \\
         \cmidrule(lr){2-3} \cmidrule(lr){4-5}
        Dataset         & \# Clients &  std      & \# Clients & std \\
        \midrule
        Caltech101      & 5          &  0.80     & 2         & 0.38 \\
        Flowers102      & 5          &  1.15     & 3         & 0.35 \\
        Oxford Pets     & 2          &  5.11     & 1         & ---  \\
        FGVC Aircraft   & 5          &  1.08     & 2         &  0.26 \\
        DTD             & 2          &  3.86     & 1         & ---  \\
        UCF101          & 5          &  1.50     & 3         &  0.66 \\
        Food101         & 5          &  0.19     & 3         &  0.11 \\
        Stanford Cars   & 10         &  0.62     & 5         &  0.40 \\
        SUN397          & 20         &  0.83     & 10        &  0.61 \\
    \midrule
        Total/Avg.      & 59        & 1.68       & 30       & 0.39\\
    \bottomrule
    \end{tabular}
\end{table}

\begin{table*}[h!]
    \centering
    \renewcommand{\arraystretch}{1.2}
    \rowcolors{2}{gray!10}{white} 
    \caption{Zero-shot classification accuracy (\%) in the homogeneous federated setting. We report only FedCoOp and FedTPG as they are the only state of the methods evaluated under this setting.}
    \label{tab:homogeneous}
    \begin{tabularx}{\textwidth}{l|*{9}{>{\centering\arraybackslash}X}|c}
    \toprule
    \rowcolor{white} 
         Method & Caltech101 & Flowers102 & Oxford Pets & FGVC Aircraft & DTD & UCF101 & Food101 & Stanford Cars & SUN397 & Avg. \\
    \midrule
           FedCoOp & 91.37 & 69.32 & \underline{96.30} & \underline{29.75} & 38.52 & 48.89 & 87.03 & 58.72 & 67.42 & 65.25 \\
           FedTPG & \textbf{93.88} & \underline{71.63} & 95.91 & 22.43 & \underline{45.29} & \underline{64.52} & \textbf{90.67} & \underline{59.47} & \underline{76.01} & \underline{68.86} \\
        \midrule
           \methnam & \underline{93.06} & \textbf{74.20} & \textbf{96.75} & \textbf{34.97} & \textbf{47.10} & \textbf{75.17} & \underline{89.98} & \textbf{74.15} & \textbf{77.50} & \textbf{73.65} \\
         
          \bottomrule
    \end{tabularx}
\end{table*}
%We finally evaluate performance in the \emph{homogeneous} scenario, where we generate a federation for each dataset and assess its generalization capabilities. As shown in Table~\ref{tab:homogeneous}, the overall performance is lower than in the heterogeneous setting, which can be attributed to the fact that the source of knowledge is now limited to a single dataset at a time. Notably, in this setting, our method significantly outperforms FedTPG (e.g., by nearly five percentage points), demonstrating that our strategy is more adaptable to new knowledge even when data source variability is restricted.

We finally evaluate performance in the \emph{homogeneous} scenario, where we generate a separate federation for each dataset. Specifically, for each dataset, we partition the available data into $C$ clients, each receiving a distinct subset of classes. As in previous evaluations, we assess zero-shot classification capabilities by using 50\% of the available classes for training and the remaining 50\% for testing. 

As shown in Table~\ref{tab:homogeneous}, the overall performance in this setting is lower than in the heterogeneous scenario, which can be attributed to the fact that the source of knowledge is now restricted to a single dataset at a time, limiting the diversity of learned representations. Despite this challenge, \methnam significantly outperforms FedTPG in most cases, achieving the highest average accuracy (73.65\%), surpassing FedTPG (68.86\%) by nearly five percentage points.

\subsection{Communication overhead analysis}
One of the key challenges in federated learning is the communication overhead incurred during training, particularly in large-scale deployments. In fully decentralized settings, where clients directly exchange model updates or prompts, reducing communication costs is critical for scalability and efficiency. In this section, we analyze the cumulative data transmitted over 500 federated rounds in a system with 59 clients, reflecting our heterogeneous scenario with  $N = 10$  classes per client. The objective of this evaluation is to compare the communication efficiency of \methnam with FedTPG (our primary competitor according to the evaluations in the previous section) and assess how different parameter configurations impact data transmission.

\begin{figure}
\centering
\includegraphics[width=0.95\linewidth]{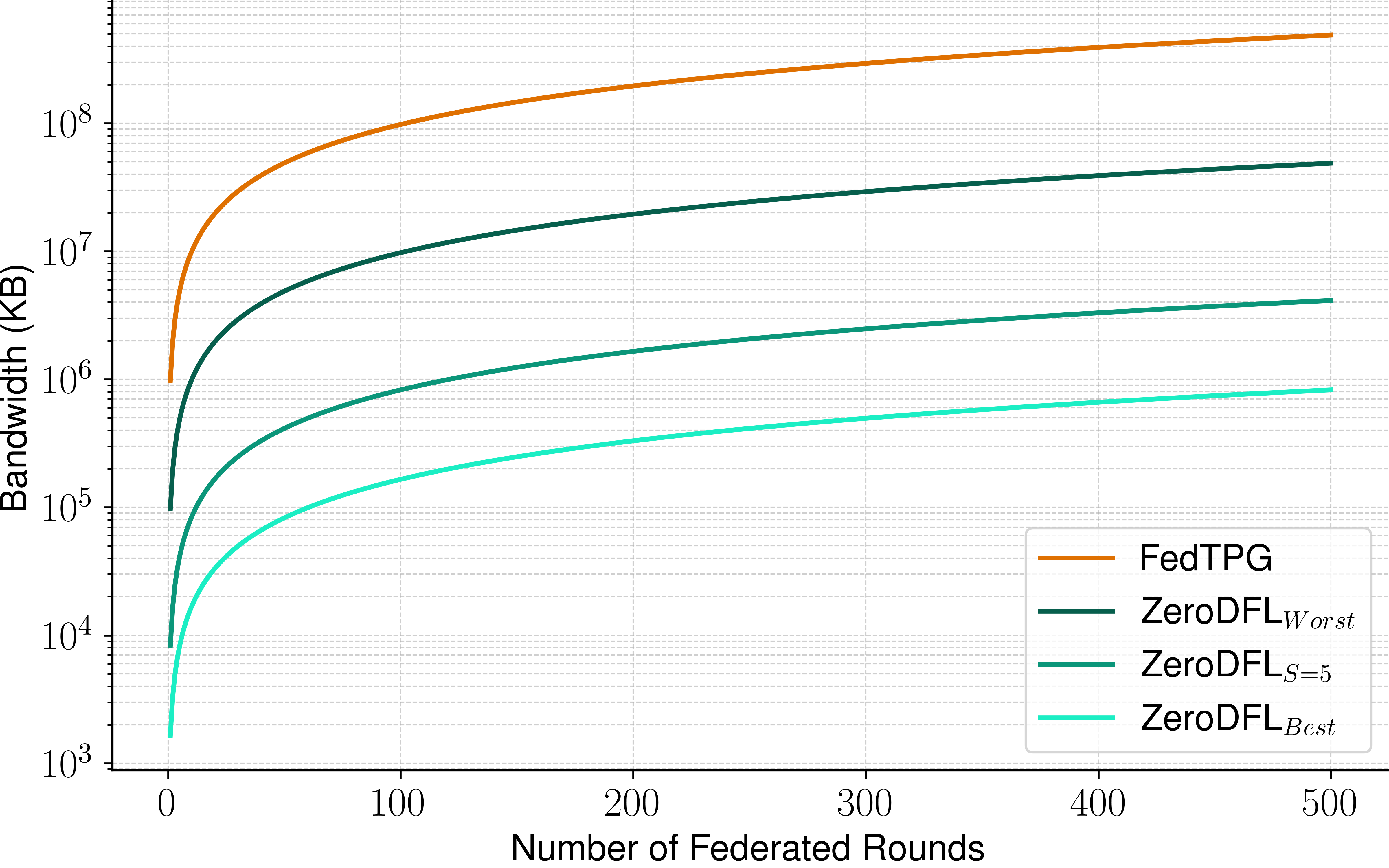}
\caption{\textbf{Cumulative communication cost} over 500 federated rounds in the heterogeneous setting with 59 clients. The plot compares FedTPG with three configurations of \methnam: \emph{Worst} (maximum communication overhead, ensuring full knowledge propagation), a balanced setting with $S=5$, and \emph{Best} (minimal communication overhead, requiring careful prompt distribution).}
\label{fig:comm_ov}
\end{figure}

Figure~\ref{fig:comm_ov} illustrates the cumulative data transmitted over 500 federated rounds. While FedTPG incurs a substantial communication overhead, accumulating approximately \cumulativefedtpg of transmitted data, this value remains fixed once the number of clients and rounds is determined. In contrast, the communication cost in \methnam is not predetermined but instead depends on several factors, such as the number of selected recipient clients per round $S$ and the number of prompt vectors per client $(M)$. Given  $M=4$ , the average amount of data transmitted by \methnam is approximately $\cumulativeourbest \times S$.

Since, in \methnam, each client independently selects its recipients, there is a nonzero probability that some clients may not receive any prompts in a given round. This issue can be mitigated by increasing $S$ , up to the extreme case where each client broadcasts its prompts to all clients in the federation. We thus analyze three different configurations to assess the trade-off between knowledge propagation and communication cost:
\begin{itemize}
    \item \textit{Worst-case scenario}: Every client broadcasts its prompts to all others $(S=C)$, ensuring maximum knowledge propagation but at the highest communication cost. This results in a total transmission of \cumulativeourworst over 500 rounds, still achieving a \reductionourworst reduction in communication overhead compared to FedTPG.
    \item \textit{$S=5$ (balanced) scenario}: Each client selects five recipients per round, maintaining a reasonable update frequency while significantly reducing communication costs. Under this setting, \methnam transmits approximately \cumulativeournormal of data over the entire training process, achieving a \reductionournormal reduction in communication overhead.
    \item \textit{Best-case (minimal communication) scenario}: Each client sends only one of its  $M$  learned prompts to  $M$  different clients, drastically reducing communication overhead to \cumulativeourbest. However, this approach introduces challenges related to ensuring an even distribution of prompts across the federation, which may impact convergence.
\end{itemize}

Such a drastic reduction in data transmission is particularly advantageous in bandwidth-constrained environments, where excessive communication overhead can be a limiting factor. Addressing the potential challenges of sparse prompt distribution lies outside the scope of this work but represents a promising direction for future research.

\begin{table*}[htb!]
    \centering
    \renewcommand{\arraystretch}{1.2}
    \rowcolors{2}{gray!10}{white} 
    \caption{Zero-shot classification accuracy (\%) in the heterogeneous setting with different numbers of exchanged prompts (\( M_s \)) per round.}
    \label{tab:prompt}
    \begin{tabularx}{\textwidth}{l|*{9}{>{\centering\arraybackslash}X}|c}
    \toprule
    \rowcolor{white}
         Shared Prompts (\( M_s \)) & Caltech101 & Flowers102 & Oxford Pets & FGVC Aircraft & DTD & UCF101 & Food101 & Stanford Cars & SUN397 & Avg. \\
    \midrule
            %Upper bound (FedCoOp) & 92.79 & 70.12 & 95.39 & 30.95 & 52.74 & 70.60 & 88.47 & 72.32 & 71.89 & 71.70 \\
    %\midrule
           4 (\methnam) & 94.70 & \textbf{78.67} & \underline{96.75} & \underline{34.94} & \underline{59.54} & \textbf{77.44} & \textbf{91.50} & \underline{75.38} & \underline{76.83} & \textbf{76.19} \\
           3 & \underline{94.75} & 78.00 & \textbf{97.37} & 33.95 & 59.05 & 76.78 & \underline{91.46} & \textbf{75.42} & \textbf{77.06} & 75.98 \\
           2 & 93.28 & \underline{78.32} & 96.14 & \textbf{35.81} & \textbf{61.95} & 76.81 & 90.96 & 74.82 & 76.63 & \underline{76.08} \\
           1 & \textbf{95.19} & 78.27 & 96.19 & 34.82 & 58.45 & \underline{77.19} & 91.31 & 74.89 & 76.09 & 75.82 \\
    %\midrule
            0 (CoOp) & 93.10 & 71.95 & 96.60 & 28.59 & 54.99 & 68.09 & 88.22 & 70.17 & 68.62 & 71.14 \\
         
          \bottomrule
    \end{tabularx}
\end{table*}

\begin{comment} 
\begin{table*}[htb!]
    \centering
    \renewcommand{\arraystretch}{1.2}
    \rowcolors{2}{gray!10}{white} 
    \caption{\textbf{Communication strategies. Accuracy (\%) on unseen clients in the heterogeneous setting with different percentages of transmitted prompts per round.}}
    \label{tab:prompt}
    \begin{tabularx}{\textwidth}{l|*{9}{>{\centering\arraybackslash}X}|c}
    \toprule
    \rowcolor{white}
         \% Exchanged & Caltech101 & Flowers102 & Oxford Pets & FGVC Aircraft & DTD & UCF101 & Food101 & Stanford Cars & SUN397 & Avg. \\
    \midrule
            %Upper bound (FedCoOp) & 92.79 & 70.12 & 95.39 & 30.95 & 52.74 & 70.60 & 88.47 & 72.32 & 71.89 & 71.70 \\
    %\midrule
           100\% (\methnam) & 94.70 & \textbf{78.67} & \underline{96.75} & \underline{34.94} & \underline{59.54} & \textbf{77.44} & \textbf{91.50} & \underline{75.38} & \underline{76.83} & \textbf{76.19} \\
           \zerogray75\% & \underline{94.75} & 78.00 & \textbf{97.37} & 33.95 & 59.05 & 76.78 & \underline{91.46} & \textbf{75.42} & \textbf{77.06} & 75.98 \\
           \zerowhite50\% & 93.28 & \underline{78.32} & 96.14 & \textbf{35.81} & \textbf{61.95} & 76.81 & 90.96 & 74.82 & 76.63 & \underline{76.08} \\
           \zerogray25\% & \textbf{95.19} & 78.27 & 96.19 & 34.82 & 58.45 & \underline{77.19} & 91.31 & 74.89 & 76.09 & 75.82 \\
    %\midrule
            \zerowhite\zerowhite0\% (CoOp) & 93.10 & 71.95 & 96.60 & 28.59 & 54.99 & 68.09 & 88.22 & 70.17 & 68.62 & 71.14 \\
         
          \bottomrule
    \end{tabularx}
\end{table*}
\end{comment}

\subsection{Communication vs. Generalization}

In this section, we analyze how varying the number of prompt vectors exchanged per rounds affects model performance in a federated learning setting. Instead of full sharing all prompts across clients, we explore a partial sharing strategy, where only a subset $M_s$ of the available $M$ prompts is exchanged with other clients in each communication round, while the remaining $M-M_s$ prompts remains local and continue to be used throughout training. This approach allow us to assess how different levels of prompt exchange influence both knowledge transfer and generalization ability in a decentralized framework. Furthermore, understanding the role of prompt exchange is crucial for optimizing communication efficiency while maintaining high classification accuracy in real-world federated applications.

%In this section, we analyze how varying the number of prompt vectors exchanged per round affects model performance in a federated learning setting. Instead of fully sharing all prompts across clients, we evaluate the impact of exchanging only a subset of the available prompts while retaining the rest locally. This enables a more detailed assessment of how different levels of prompt exchange influence both knowledge transfer and generalization ability in a decentralized framework. Furthermore, understanding the role of prompt exchange is crucial for optimizing communication efficiency while maintaining high classification accuracy in real-world federated applications.

By default, each client (in the current settings) is given $M = 4$ learnable prompt vectors. Table~\ref{tab:prompt} presents the results for different values of $M_s$, ranging from 0 (local-only learning, no exchange) to 4 (full prompt exchange). Lower values of $M_s$ reduce communication overhead, which is particularly advantageous in bandwidth-constrained environments and real-world scenarios where frequent data exchanges may be impractical. However, lower exchange rates may limit the richness of shared knowledge across clients, as fewer prompts are available for collective adaptation. Conversely, higher values of $M_s$ enable better knowledge transfer, as more comprehensive prompt sharing allows clients to benefit from global insights at the cost of increased communication. This trade-off between communication efficiency and model performance is a fundamental aspect of federated learning, particularly in decentralized systems with varying network constraints.

%By default, each client (in the current settings) maintains $M = 4$ learnable prompt vectors. Table~\ref{tab:prompt} presents the results for different values of $M$, ranging from 0 (local-only learning) to 4 (full exchange). Lower values of $M$ reduce communication overhead, which is particularly advantageous in bandwidth-constrained environments and real-world scenarios where frequent data exchanges may be impractical. However, lower exchange rates may limit the richness of shared knowledge across clients, as fewer prompts are available for collective adaptation. Conversely, higher values of $M$ enable better knowledge transfer, as more comprehensive prompt sharing allows clients to benefit from global insights at the cost of increased communication. This trade-off between communication efficiency and model performance is a fundamental aspect of federated learning, particularly in decentralized systems with varying network constraints.

Our findings confirm that exchanging all $M$ prompts enhances generalization, as it facilitates better knowledge propagation across clients. This result underscores the importance of prompt exchange in federated learning, allowing individual clients to leverage collective insights while maintaining decentralized training. Notably, even partial prompt exchange leads to substantial performance improvements over isolated local learning (CoOp), demonstrating the potential for optimized prompt-sharing configurations that strike a balance between communication efficiency and model accuracy. These results indicate that even a fraction of exchanged prompts can significantly contribute to model generalization, reducing the need for excessive communication overhead.

Interestingly, in some cases, exchanging fewer prompts leads to better performance on specific datasets. This counterintuitive effect may be attributed to higher intra-class differentiation within certain datasets, where local fine-tuning plays a more critical role in capturing dataset-specific nuances. Specifically, when datasets contain highly domain-specific classes, retaining a portion of locally fine-tuned prompts can act as a form of personalization, allowing the model to develop a richer and more specialized representation of class variations without excessive reliance on global prompt aggregation. This observation suggests that over-sharing prompts may sometimes dilute personalized features, emphasizing the importance of adapting prompt exchange levels based on dataset properties.

Overall, these findings highlight the need for adaptive prompt-sharing mechanisms, where the number of exchanged prompts is dynamically adjusted based on dataset properties, task complexity, and communication constraints. 
 % All methods were tested in a Non-IID setting where the classes of the train data were distributed to multiple clients. Each client owns \textit{n} = 20 completely disjoint classes where each class has eight labeled images for few-shot training. Two training strategies were adopted: in the first, all datasets were used simultaneously, i.e. there were several clients with different datasets; in the second, the datasets were selected individually. 

 %In addition, the average performance of each method across all datasets was also evaluated. 

\section{Conclusion}
In this work, we introduced \methnam, a fully decentralized approach for federated prompt learning, which eliminates the need for a central server while ensuring efficient knowledge propagation and providing high generalization capability to each client. Our method leverages an iterative process of adaptation and exchange, where clients directly share optimized prompt vectors with one another. Through a weighted selection mechanism, clients prioritize less frequently selected nodes, ensuring a balanced distribution of knowledge and improving generalization across unseen classes.

Our empirical analysis demonstrates that \methnam achieves state-of-the-art performance while significantly reducing communication overhead. Compared to centralized approaches, our framework achieves up to a \reductionournormal reduction in transmitted data, making it particularly suitable for bandwidth-constrained environments. This reduction in communication cost is particularly valuable in real-world federated learning deployments, where devices operate under limited network resources and excessive data transmission can be prohibitive. Additionally, the decentralized nature of our method improves resilience against single points of failure, making it more robust in dynamic and large-scale federated systems.

Beyond efficiency, our work enhances privacy and security in federated learning by leveraging decentralization and text-based prompt sharing. By eliminating the central aggregator, \methnam removes a single point of failure and reduces the risk of centralized attacks. In addition, sharing only text-based prompts inherently limits the exposure of sensitive information. Since prompts do not contain raw data or feature embeddings, they significantly reduce the attack surface and mitigate information leakage risks.
Furthermore, the fully decentralized structure enhances resilience against targeted attacks~\cite{NIU2024404}, preventing adversarial manipulation and improving system robustness. This makes \methnam particularly well-suited for security-critical applications such as healthcare, finance, and edge AI, where both data privacy and robustness are essential. At the same time, it maintains strong adaptability to heterogeneous, non-IID data distributions without compromising security.

In future work, we plan to further optimize the communication strategy by investigating how to maintain a specialized subset of prompts locally at each client while exchanging more general prompts throughout the federation. This approach could refine the balance of knowledge distribution, enhancing both efficiency and adaptability in decentralized federated prompt learning. Additionally, we aim to explore dynamic prompt-sharing mechanisms, where the number of exchanged prompts is automatically adjusted based on dataset properties, learning objectives, and communication constraints. Such an adaptive framework could further optimize model performance while maintaining low communication costs.

%The broader impact of our work extends to real-world federated AI applications, such as privacy-preserving medical imaging, distributed autonomous systems, and on-device learning for edge AI applications. By providing a scalable, communication-efficient, and privacy-friendly solution, \methnam paves the way for future advancements in decentralized federated learning, pushing the boundaries of efficient and secure collaborative AI.

%Future work will explore additional optimizations to the communication strategy, such as dynamically adjusting the number of selected recipients per round or investigating more advanced sampling techniques for prompt selection. 

\section{Acknowledgments}
The authors acknowledge financial support from PNRR MUR project PE0000013-FAIR.
\bibliographystyle{IEEEtran}
\bibliography{egbib}

\end{document}